\documentclass[10pt,journal,compsoc]{IEEEtran}
\ifCLASSOPTIONcompsoc
  \usepackage[nocompress]{cite}
\else
  \usepackage{cite}
\fi
\usepackage[ruled,vlined]{algorithm2e}

\ifCLASSINFOpdf
   \usepackage[pdftex]{graphicx}
   \graphicspath{{../pdf/}{../jpeg/}}
  \DeclareGraphicsExtensions{.pdf,.jpeg,.png,.jpg}
\else
\fi
\usepackage{algorithmic}
\usepackage{verbatim}

\begin{document}
\title{TempNodeEmb:Temporal Node Embedding considering temporal edge influence matrix}

\author{Khushnood~Abbas,~\IEEEmembership{Member,~ACM,}
        Alireza~Abbasi,%~\IEEEmembership{Fellow,~OSA,}
        Dong Shi,
        Niu Ling,
        Mingsheng Shang,
        Chen Liong,
        and~Bolun Chen.% <-this % stops a space
\IEEEcompsocitemizethanks{\IEEEcompsocthanksitem Khushnood Abbas, Shi dong and Niu Ling are with the School of Computer Science, Zhoukou Normal University, Henan, China. Alireza Abbasi is with School of Engineering and IT, University of New South Wales, Canberra Australia. Mingsheng Shang is with Chongqing Institute of Green and Intelligent Technology, Chinese Academy of Sciences, China. Liyong Chen is with School of Network Engineerig Zhoukou Normal Univeristy. Bolun Chen is with Huayin Institute of Technology, Huaian, Jiangsu, China.
\protect
E-mail: Khushnood.abbas@acm.org
%\IEEEcompsocthanksitem J. Doe and J. Doe are with Anonymous University.
}% <-this % stops an unwanted space
\thanks{Manuscript received Feb 19, 2020; revised March 26, 2020.}}


\markboth{Journal of \LaTeX\ Class Files,~Vol.~14, No.~8, August~2020}%
{Abbas \MakeLowercase{\textit{et al.}}: IEEE journal}



\IEEEtitleabstractindextext{%
\begin{abstract}
Understanding the evolutionary patterns of real-world evolving complex systems such as human interactions, transport networks, biological interactions, and computer networks has important implications in our daily lives. Predicting future links among the nodes in such networks reveals an important aspect of the evolution of temporal networks. To analyse networks, they are mapped to adjacency matrices, however, a single adjacency matrix cannot represent complex relationships (e.g. temporal pattern), and therefore, some approaches consider a simplified representation of temporal networks but in high-dimensional and generally sparse matrices. As a result, adjacency matrices cannot be directly used by machine learning models for making network or node level predictions. To overcome this problem, automated frameworks are proposed for learning low-dimensional vectors for nodes or edges, as state-of-the-art techniques in predicting temporal patterns in networks such as link prediction. However, these models fail to consider temporal dimensions of the networks. This gap motivated us to propose in this research a new node embedding technique which exploits the evolving nature of the networks considering a simple three-layer graph neural network at each time step, and extracting node orientation by Given’s angle method. To prove our proposed algorithm’s efficiency, we evaluated the efficiency of our proposed algorithm against six state-of-the-art benchmark network embedding models, on four real temporal networks data, and the results show  our model outperforms other methods in predicting future links in temporal networks.
\end{abstract}
}
\maketitle

\IEEEdisplaynontitleabstractindextext

% \fi
\IEEEpeerreviewmaketitle

\IEEEraisesectionheading{\section{Introduction}\label{sec:introduction}}

\IEEEPARstart {T}{emporal} graphs are amongst the best tools to model real-world evolving complex systems such as human interactions, transport networks, the Internet, biological interactions, and scientific networks, to name a few. Understanding the evolving patterns of such networks have important implications in our daily life and predicting future links among the nodes in networks reveals an important aspect of the evolution of temporal networks \cite{Abbasi_2016}. Learning useful representations from networks (or graphs) not only reduces the computational complexity but also provides greater predictive power that facilitates the use of machine learning methods \cite{wang2017knowledge}. To apply mathematical models, networks are represented by adjacency matrices in which only local information of each node is considered are both high-dimensional and generally sparse in nature, and therefore, insufficient for representing global information which are often important features of the network such as nodes' neighbours information. As a result, it cannot be directly used by machine learning models for predicting graph or node level changes. Similarly representing temporal networks using temporal adjacency matrices, as snapshot of the network at different time steps, inherits the same problems and necessitates using alternatives methods. This has led to development of deep neural network based approaches to learn node/edge level features \cite{bengio2013representation}. \par 

This research presents a new deep learning based model for generating low-dimensional features from large high-dimensional networks considering their temporal information. Our technical contributions are as follows:

 \begin{enumerate}
 \item We considered time varying adjacency matrix whose entries are ${e_{i,j,t}} = {e^{t - {t_{now}}}}$, where $t$ is the time step when graph was constructed, and $t_{now}$ is the current time.
 \item We developed a simple three-layer Graph convolutional feed forward model without implementing nonlinear activation and parameter learning, instead of a complex static generating method.
 \item We considered angles (using Given's angle method) between any two consecutive time steps, calculated from static generated features, and solved the least square optimization problem using QR factorization method.
 \end{enumerate}
The remainder of the paper is organised as follows. We reviewed some related works in the direction of node embedding. Then we formally defined the problem in Section $3$. In Section $4$ we present our proposed approach for embedding temporal networks, which we refer to as $TempNodeEmb$. We outline our experimental design including data sets, evaluation metrics and benchmark methods in Section $5$, and present the results in Section $6$. We close the paper in Section $7$ with a discussion and our conclusions .

\section{Related Works}
Currently deep learning based framework is found to be very effective in learning low-dimensional representation for euclidean data such as image video or audio, meaning these data sets can be easily represented in the form of grid structure without loss of information.  Network embedding, as such an approach, is developed for learning hidden representations of nodes in a network to encode links in a continuous vector space to facilitate the use of statistical models \cite{perozzi2014deepwalk}. In other words, very large high-dimensional and sparse networks embeds into low-dimensional vectors \cite{tang2015line}, while integrating global structure of the network (maintaining the neighbourhood information) into the learning process \cite{cao2015grarep}, that has applications in tasks such as node classiﬁcation, visualization, link prediction, and recommendation \cite{tang2015line, mahdavi2018dynnode2vec}.
Although network embedding models are best to capture network structural information, they lack considering temporal granularity and fail in temporal level predictions such as temporal link prediction, and evolving communities prediction. One solution for generating node embedding in temporal network is to generate static embedding for each time step and find node level orientation based on previously generated features \cite{haddad2019temporalnode2vec, singer2019node}.

Network evolution studies have been at the center of network studies \cite{leskovec2005graphs, Abbas2018Popularity, yu2016network, albert2002statistical, trivedi2017know} and in particular on addressing link prediction \cite{lu2011link}. Apart from traditional machine learning and statistical modeling approaches currently deep neural networks are also being developed \cite{cui2018survey}. These models effectively generate a $d$ (lower than total number of nodes in graph) dimensional feature vector based on graph structure, which can be fed into any machine learning model. Some researchers came up with matrix factorization approach \cite{cao2015grarep,ou2016asymmetric}. Further researchers have used deep neural networks auto-encoders \cite{cao2016deep,wang2016structural}, convolutional neural networks\cite{kipf2016semi} often considering random walks \cite{chen2018harp,perozzi2014deepwalk,grover2016node2vec}. All these methods considered only static nature of the graph, and only recently some researchers considered temporal aspect of the network for low-dimensional node feature embedding \cite{mahdavi2018dynnode2vec, NguyenContinousTime2018, peng2019dynamic,li2018deep}. Methods for embedding network temporal behaviour are developed and among them is applying a static method at every time-step and minimizing error based on consecutive time step embedding \cite{singer2019node}.

\section{Problem definition}
Graphs are the best choice in representing implicit or explicit relationship among entities and are recently emerged as one of the best data structure to store such heterogeneity in data. Graphs are composed of a set of nodes $V = \{ {v_1},{v_2}...,{v_{\left| V \right|}}\}$ and a set of edges ${E_{}} = {e_{i,j}}$ that reflect a connection between each pair of nodes. However, when we model the real interactions in our daily life, the associated edges ${E_T} = {e_{i,j,t}}$ contains a time stamp $t$, where $i,j,t$ represents an interaction between node $v_i$ and $v_j$ at time $t$. So, a dynamic or temporal graph $G$ can be represented by a three tuple set $G(V,E_T)$: $G_t$, the graph at time $t$, contains all the edges which has been formed before time $t$. For training our model, we considered $T$ time slices such that $t \in [1,T]$. Consequently we use $T$ set of temporal graphs ${G_1},{G_2}....,{G_T}$. So, we aim to predict if an edge will be formed between two nodes $v_i$ and $v_j$ at time $T+t'$.\par

\begin{figure*}[!t]
\centerline{\includegraphics[width=0.9\textwidth]{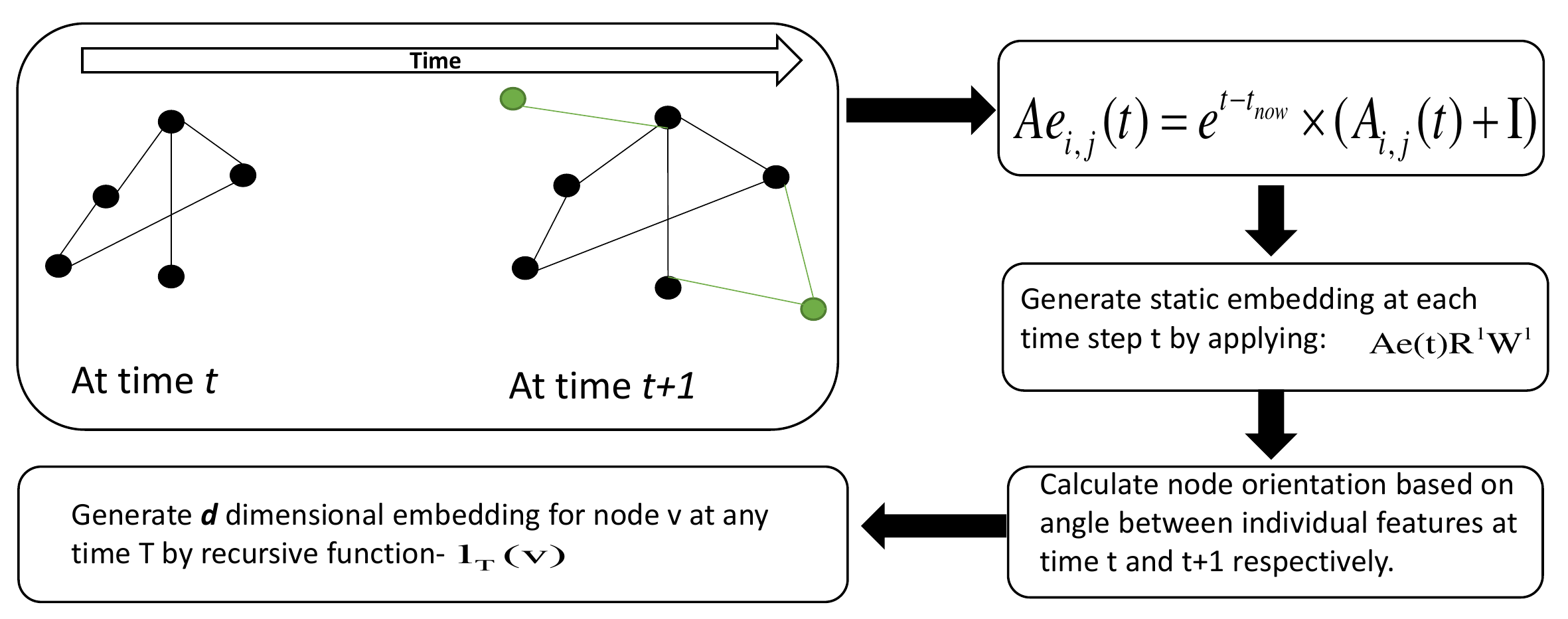}}

  \caption{This is the proposed model framework to generate $d$ dimensional node embedding for temporal graphs.}
\label{fig:modelFramework}
\end{figure*}

\section{Method}
In this section we describe our proposed method and its main contributions. In Fig. \ref{fig:modelFramework}, we show our proposed framework that includes the following steps:
\begin{comment}
\subsection{Graph convolution operation}
At every time step $t$ from training set, we generate a $d$ (${\rm{d  <   <   <  |V|}}$, $|V|$ is the number of nodes in the graph) dimensional feature vector for every node, by applying the following convolution operation. 
%For a given graph $G(V,E)$, at any time $t$ the timed adjacency matrix is represented as $A_{i,j}(t)$, and the temporal edge matrix as ($A{e_{i,j}}(t)$) can be formulated as:
%\begin{equation}
%A{e_{i,j}}(t) = {e^{t - {t_{now}}}} \times ({A_{i,j}}(t) + I)
%\end{equation}\\
%--------------------------------------------------
For a given graph $G(V,E)$, at time $t$ we apply the Graph Convolution as follows:
\begin{enumerate}
\item the timed adjacency matrix is represented as $A_t$ at time $t$ with size $|V|\times |V|$. We add self-loop by adding identity matrix $I; $ ${{\hat A}_t} = {A_t} + I$, where only diagonal elements are $1$ and rest elements are $0$ (represents only self-loops: node $i$ links to itself).
\item Further we create a temporal edge influence matrix: $A_{i,j}(t)$.
\item Suppose every node has its own feature vector of size $F^0$, there will be a feature matrix of size ($|V|\times F^0$).
\item Using temporal edge influence matrix and node feature matrix, we apply simple convolution operations without non-linear activation operation. \\
\end{enumerate}
\\
\end{comment}
\subsection{Temporal edge influence matrix}
To consider temporal influence of nodes, we considered that edge influence of a node decreases exponentially. Suppose time adjacency matrix at any time $t$ is represented by $A_{i,j}(t)$, the temporal influence matrix ($A{e_{i,j}}(t)$) at any time can be formulated as:
\begin{equation}
A{e_{i,j}}(t) = {e^{t - {t_{now}}}} \times ({A_{i,j}}(t) + I)
\end{equation}
where, $I$ is identity matrix, as above. \\

\subsection{Generating Static Embedding}
At every time step, we generate a static $d$-dimensional embedding for every node $v$, using a three-layer of convolutional neural network. We generate a static embedding matrix $\Theta _t$ at every time step $t$, which we use the simplest convolutional forward prorogation model as follows:
\begin{equation}
f({R^l},A) = {{\hat A}_{et}}{R^l}{W^l}
\end{equation}
where $R^l$ is hidden representation and $W^l$ is a random weight matrix at layer $l$, and $R^0 = I$; as we do not have node level features at time 0.
Two things to notice here is that, we do not apply either the degree matrix normalization technique \cite{kipf2016semi} nor nonlinear activation. As we are generating embedding without supervised learning. Considering the non-linear activation was not giving us good results.\\

\subsection{Loss function and learning representation}
Our aim is to learn feature vector at time step $T$ using function ${{\rm{l}}_{\rm{T}}}{\rm{(v)}}$. For temporal link prediction tasks, we learn the parameter using cross entropy loss, as follows:
\begin{equation}
%\begin{gathered}
  Cost(p,\widehat p) =  - p\log (\widehat p) - (1 - p)\log (1 - \widehat {{p}})
%\end{gathered}
\label{binaryCrossEntropy}
\end{equation}
where $p$ is the actual label and $\widehat p$ is the predicted label.
%In our link prediction problem we have considered function  $C$  as concatenation function  between features of node $v_1$ and node $v_2$.
As link prediction tasks happens between two nodes, we used concatenation function. Although it can be any function according to requirements. Further we learn the  ${{\rm{l}}_{\rm{T}}}{\rm{(v)}}$ as follows:
\begin{equation}
{{\rm{l}}_{\rm{T}}}{\rm{(v) = }}{{\rm{L}}_T}(v,{G_1},{G_2},....,{G_T})
 \end{equation}
Further we learn the final orientation using recursive function, described by Singer et al. \cite{singer2019node} as follows:
\begin{equation}
 {{\rm{l}}_{\rm{T}}}({\rm{v}}){\rm{ = }}\sigma ({A_{et}}{l_t}(v) + B{\Theta _t}{\Phi _t}v)
\end{equation}
 where ${A_{te}},B,{\Theta _t},{\Phi _t}$ are matrices which are learned during training and $\sigma $ is activation function. In our case, we use $\tanh$ function, where $\Theta _t$ is the static feature matrix at time $t$. %We found node orientation by calculating angle between individual features at consecutive time step $t$ and $t+1$ as follows.
\break
\subsection{Calculating node alignment}
One of the important tasks in modeling temporal networks embedding is finding node alignments over time. In this work, we calculate the angle between node vectors (feature spaces). Instead of calculating the angles between two nodes, we calculate how node's individual features are changing. As two features, at times $t$ and $t+1$, lie in the same euclidean space, we consider the angle between features at two different time steps as described by angles between two scalars \cite{demmel1990matrix}. \break 
Using the two adjacency matrices of a graph at times $t$ and $t+1$, we create the angle between its individual features using Algo. \ref{algo:calculatingAngles}. In order to know how each feature aligns over time, we create matrices $\Theta _{\cos \alpha }$ and $\Theta _{\cos \beta }$. Furthermore, we apply dot operations so as to find the angle orientation with respect to the other nodes i.e. matrix $C_t = {\Theta ^T}_{\scriptstyle\cos \beta \hfill\atop
\scriptstyle\hfill} \times {\Theta _{\cos \alpha }}$. 
To find a stable matrix between any two consecutive snapshots, we find orthogonal matrix $Q_t$ using the decomposition method $i.e. C_t = Q_t * R_t$ (i.e. we used QR decomposition method). By using recursive function, we find stable basis column matrix of $Q_t$ (at least first $k$ columns will be ortho-normal basis) which can be used to project snapshot of network at $t$ to future time $t+1$.
%%%%

%%%%%%%%%%
\begin{algorithm}
\caption{Calculating the angles($x_{v}(t,i)$),$x_{v}(t+1,i)$}
\begin{algorithmic}
%\REQUIRE $n \geq 0 \vee x \neq 0$
\IF{$x_{v}(t+1)=0$}
\STATE $cos\alpha=1;cos\beta=0$
\ELSE
\IF{$|x_{v}(t+1,i)| > |x_{v}(t,i)|$}
\STATE $tmp=-x_{v}(t,i)/x_{v}(t+1,i)$
\STATE$cos\beta=1/\sqrt {1 + tm{p^2}}$
\STATE$cos\alpha=tmp.cos\beta$
\ELSE
\STATE $tmp=-x_{v}(t+1,i)/x_{v}(t,i)$
\STATE$cos\alpha=1/\sqrt {1 + tm{p^2}}$
\STATE$cos\beta=tmp.cos\alpha$
\ENDIF
\ENDIF
\\
\Return $cos\alpha, cos\beta$
\end{algorithmic}
\label{algo:calculatingAngles}
\end{algorithm}

\subsection{Learning for link prediction}
After getting $d$ dimensional stable aligned vector for each node and each time, we use recurrent neural network and in particular the Gated Recurrent Units \cite{cho2014learning}, which is a new kind of gating mechanism that has two gates, namely reset and update, and fewer parameters than 'long short-term memory' (LSTM) as an artificial recurrent neural network architecture. This way, we train the network by formulating our link prediction problem as a binary classification problem. Further we concatenate the features of any two nodes so that neural network can learn probability scores of having a link between any two nodes. We use binary cross entropy loss as in equation \ref{binaryCrossEntropy}.

\subsection{Optimization algorithm}
For parameter learning, we use Adam optimizer \cite{kingma2014adam}, which calculates an exponentially weighted average of past gradients and removes biases ($v^{corrected}$). Further, it calculates the exponentially weighted average of the squares of the past gradients and removes biases($s^{corrected}$). The details are as follows:
\begin{equation}
{v_{d{W^{[l]}}}} = {\beta _1}{v_{d{W^{[l]}}}} + (1 - {\beta _1})d{W^{[l]}}v_{d{W^{[l]}}}^{corrected} = \frac{{{v_{d{W^{[l]}}}}}}{{1 - {{({\beta _1})}^t}}}
\end{equation}
\begin{equation}
{s_{d{W^{[l]}}}} = {\beta _2}{s_{d{W^{[l]}}}} + (1 - {\beta _2}){(d{W^{[l]}})^2}s_{d{W^{[l]}}}^{corrected} = \frac{{{s_{d{W^{[l]}}}}}}{{1 - {{({\beta _1})}^t}}}
\end{equation}
\begin{equation}
W^{[l]} = W^{[l]} - \alpha \frac{v^{corrected}_{dW^{[l]}}}{\sqrt{s^{corrected}_{dW^{[l]}}} + \varepsilon}
\end{equation}
\begin{equation}
b^{[l]} = b^{[l]} - \alpha \frac{v^{corrected}_{db^{[l]}}}{\sqrt{s^{corrected}_{db^{[l]}}} + \varepsilon}
\end{equation}

where, $t$ counts the number of steps taken of Adam, $l$ is the number of layers ($1,2,3$), $\beta_1$ and $\beta_2$ are hyper-parameters that control the two exponentially weighted averages. $\alpha$ is the learning rate. $\varepsilon$ is a very small number to avoid dividing by zero. Following the same aforementioned steps, parameter $b^{[l]}$ is also updated for each layer $l$. \\

\section{Experimental Design}
In this section, we discuss our approach to evaluate our proposed method compared to relevant approaches on temporal link prediction using node embedding. Temporal link prediction problem is to predict if any two nodes will be connected by t+1 or not. To achieve this, we divide the temporal network data sets into two parts based on a pivot time, considering $70\%$ edges remains for training and the remaining $30\%$ for testing purpose. So all the edges formed at or before the pivot time are considered positive examples for training set. All the edges formed after pivot time and before future test time point is considered as positive test example for testing. Almost similar number of edges are randomly sampled to create negative examples. For negative test examples, we sample similar number to the positive test examples which have not been formed at any time. \par

\subsection{Baseline Methods}
To evaluate the performance of our proposed model, we compared it with state-of-the-art temporal embedding as well static node embedding methods, as follows:
\begin{enumerate}
\item \textbf{tNodeEmbed \cite{singer2019node} }: This method is state-of-the art for node embedding for dynamic graphs. It learns embedding by first generating static embedding and then finding node alignment. Further it can fed to recurrent neural network for task oriented predictions. 
\item\textbf{Prone \cite{ijcai2019ProNE}}: This method first initializes the embedding using sparse matrix factorization and spectral analysis for local and global structural information. \\
\item\textbf{DeepWalk\cite{perozzi2014deepwalk}}: This model learns node low dimensional embedding based on random walks. It has two hyper-parameter: walk length $l$, and window size $w$.  \\
\item\textbf{Node2vec \cite{grover2016node2vec}}: It is a similar model for graphs which works on similar principal of Word2vec model \cite{mikolov2013efficient}, as a state-of-the-art framework for word embedding in natural language processing. Based on similar skip-gram concept node2vec works on neighbourhood node and generates low-dimensional embedding. Node2vec can be generalized according to need such as if one wants to embedd similarity based on distance or based on role of the node in network. \\
\item\textbf{LINE \cite{tang2015line}}: This model generate node low level embedding considering first order and second order node similarity. Furthermore, this model uses sampling based on edge weights which also improves the performance for large scale networks. It is special case of DeepWalk when the size of vertices context is kept $1$. \\
\item\textbf{Hope \cite{ou2016asymmetric}}: The High-Order Proximity preserved Embedding method is based on PageRank and Katz index. This method uses singular value decomposition for making low rank approximations.   \par
\end{enumerate}

\subsection{Evaluation metrics}
\textbf{ROAUC}: The $ROAUC$ (aka $ROC$) is area under plot between True Positive Rate ($TPR$) and False Positive Rate ($FPR$). It depicts the trade-off between true positive and false positive prediction rate. The $TPR$ is also known as sensitivity, recall or probability of detection. ROAUC gives measure of separability of the classifier, therefore it is a vital metrics in our case also. \\
\textbf{PRAUC}: The PRAUC is the area under Precision and Recall curve, which is used to estimate the accuracy between precision and recall both at one time. In other words, precision recall pair points are obtained by considering different threshold values. Therefore, it is used to estimate efficiency for imbalance class and proves model's ability to work for skewed distributed datasets.   \par 

\subsection{Datasets}
The following real-world datasets are used in our experiments: 
\begin{enumerate}
\item \textbf{MIT human contact (MITC) network} (from \cite{realityMiningMIT2015Download}). This undirected network contains human contact data among students of the Massachusetts Institute of Technology (MIT), collected by the Reality Mining experiment performed in $2004$ as part of the Reality Commons project \cite{realityMiningMIT2015}. A node represents a person and an edge indicates the corresponding nodes have physical contacts. The data was collected over $9$ months using mobile phones. A daily granularity is used for time steps in this dataset. \\

\item \textbf{College text message (COLLMsg) network}. Data was gathered though a Facebook like social networking app used at University of California Irvine. Nodes are people and a directed edge represent a message has send from one user to another. A daily granularity between $15th$ April $2004$ to $26th$ October $2004$ is used for time steps. \\

\item \textbf{Protein-Protein interaction (PPI) network} (from \cite{singer2019node}). It includes protein as nodes and edges between any pair of proteins exist if they are biologically interacted. We consider the interaction discovery dates as the edge’s timestamp.  \\

\item \textbf{Wikipedia news edits (WikiEdit) network} (from \cite{nr-aaai15}). This is a dataset for user edits on wiki news. A monthly granularity is used for time steps in this dataset. \\

\end{enumerate}
Table \ref{tab:datasetProperties} provides basic properties of datasets including the number of nodes, links and the average degree of networks.

\begin{table}
 \centering
 \def\~{\hphantom{0}}
 \begin{minipage}{130mm}
  \caption{Properties of the dataset used in our experiments}
\label{tab:datasetProperties}
  \begin{tabular*}{20pc}{c|c|c|c}
  %{@{}l@{\extracolsep{\fill}}c@{\extracolsep{\fill}}r@{\extracolsep{\fill}}r@{\extracolsep{\fill}}r@{\extracolsep{\fill}}r@{\extracolsep{\fill}}l@{\extracolsep{\fill}}c@{\extracolsep{\fill}}c@{\extracolsep{\fill}}c@{}}
  \hline
             & \textbf{No of Nodes} & \textbf{No of edges} & \textbf{Average degree} \\

 \hline \textbf{MITC} & 96                       & 2539                 & 52.8958                 \\
\hline \textbf{COLLMsg} & 1,899                     & 59835                & 14.5729                 \\
\hline \textbf{PPI} & 16,458                    & 144,033               & 17.5031                 \\
 \hline \textbf{WikiEdit} & 25042                    & 68678                & 5.485
\end{tabular*}
\end{minipage}
\vspace*{-6pt}
\end{table}

\section{Results}
The performance of our proposed link prediction models ($TempNodeEmb_{Static}$ and $TempNodeEmb$) compared to baselines models on four real datasets are reported in Table \ref{aucPerformanceTable}. As shown, our temporal proposed model ($TempNodeEmb$) outperforms all the models and on all the data sets except in two cases. Deepwalk method performs slightly better than $TempNodeEmb$ on MIT Human contact network (MITC) considering both evaluations metrics, while its performance is not on top list among other models for the other three datasets. The small size of MITC dataset might be a key factor on the high performance of Deepwalk model, in another case, tNodeEmbed performs better on WikiEdit network considering ROC metric, while again tNodeEmbed performance in rest of the cases is not as good as Prone and Hope.
Here $TempNodeEmb_{Static}$ is a model we used to generate the static embedding at each temporal point during training process. We also want to emphasis that as Graph Convolutional model uses node level explicit features, and therefore, our model can consider node level features along with network structural features. Due to absence of node level features, we used only one hot vector for each node. We believe our model's accuracy will improve when we use node level explicit features along with temporal and graph level features.

\begin{table*}[!t]
 \centering
 \def\~{\hphantom{0}}
 \begin{minipage}{130mm}
  \caption{Comparing performance of link prediction models (rows) on four real datasets (columns) considering  ROC-AUC(ROC) and Precision-AUC(PRAUC) evaluation metrics}
\label{aucPerformanceTable}
  \begin{tabular*}{30.5pc} {|c | c c | c c  | c c  | c c | }
%  \hlineB{2}
    \bf Datasets & \multicolumn{2}{l}{\qquad  \bf PPI} &\multicolumn{2}{l}{\qquad \bf COLLMsg}  & \multicolumn{2}{l}{\qquad \bf MITC} & \multicolumn{2}{l}{\qquad \bf WikiEdit}  \\
  \hline
  \bf Models  &\bf ROC &\bf PRAUC&\bf ROC &\bf PRAUC&\bf ROC &\bf PRAUC&\bf ROC & \bf PRAUC \\
\hline
\bf Prone  & 0.756 & 0.766 & 0.620  & 0.606 & 0.692 & 0.668 & 0.782 & 0.785 \\
\hline
\bf Node2vec & 0.693 & 0.688 & 0.533 & 0.530 & 0.588 & 0.582 & 0.717 & 0.714 \\
\hline
\bf Line    & 0.700  & 0.697  & 0.559 & 0.533 & 0.618 & 0.606 & 0.690 & 0.682 \\
\hline
\bf Hope    & 0.787 & 0.791  & 0.640 & 0.621 & 0.677 & 0.646  & 0.741  & 0.761 \\
\hline
\bf Deepwalk  & 0.690 & 0.684 & 0.525 & 0.514 &  \bf 0.706 &  \bf 0.675 & 0.720 & 0.732   \\
\hline
\bf  $TempNodeEmb_{Static}$  & 0.687  & 0.685 & 0.613 & 0.662 & 0.553 & 0.555 & 0.691 & 0.694 \\
\hline
\bf tNodeEmbed    & 0.759 & 0.766 & 0.602 & 0.600 & 0.640 & 0.607 & \bf 0.798 & 0.774\\
\hline
\bf $TempNodeEmb$  & \bf 0.818 &\bf  0.821  & \bf 0.776 & \bf 0.762 & 0.690 &  0.652  & 0.782 & \bf  0.798 \\
\hline
\end{tabular*}
\end{minipage}
\vspace*{-6pt}
\end{table*}

\section{Conclusion}
Temporal node embedding has just gaining attention recently. As it provide not only node level task such as link prediction, node classification or anomaly detection but can also be applied for graph level prediction tasks such as community detection. In this work we presented an model to generate node embedding in temporal graphs. The embedding we present is very simple and effective. To achieve this, we created a temporal influence matrix and generated static embedding of nodes at each time step, applying basic 3-step forward graph convolutional operations. The only difference here is that we did not use degree matrix normalization trick as our matrices are already normalized, i.e. its entries lie between $[0,1]$ due to exponential decay operator. The second most important difference is that we did not apply any non-linear activation function. Just only 3-step convolutional operations makes performance efficient as compared to tNodeEmbed model which is based on Node2Vec. Furthermore, tNodeEmbed method does not allow node level explicit feature consideration, while our model allows node level explicit feature consideration. \\
In this work, as the node level features are not available, we therefore initialized features as one-hot vectors. To evaluate the performance of our temporal link prediction model, we used five static and one dynamic link prediction models as benchmarks. Although our current work proves to be better than state-of-the art but in future we will try to improve its performance further, as learning static feature vector and alignment at each time-step is computational in-efficient as compared to models for static graphs.

% or


\ifCLASSOPTIONcompsoc
  
  \section*{Acknowledgments}
\else
  \section*{Acknowledgment}
\fi
This work is supported in part by the Key Scientific and Technological Research Projects in Henan Province under Grants ($202102210379$,$182102210152$, $182102310034$), Zhoukou Normal University super scientific project $ZKNUC2018019$, Key scientific research projects of Henan Provincial Department of Education $20A520046$, Chinese National Natural Science Foundation under grant No. $61602202$, the Natural Science Foundation of Jiangsu Province under contracts $BK20160428$, the Six talent peaks project in Jiangsu Province under contract $XYDXX-034$ and the project in Jiangsu Association for science and technology.

\ifCLASSOPTIONcaptionsoff
  \newpage
\fi

 

%\bibliographystyle{IEEEtran}
\bibliographystyle{IEEEtran}
\bibliography{sigproc}

\begin{thebibliography}{10}
\providecommand{\url}[1]{#1}
\csname url@samestyle\endcsname
\providecommand{\newblock}{\relax}
\providecommand{\bibinfo}[2]{#2}
\providecommand{\BIBentrySTDinterwordspacing}{\spaceskip=0pt\relax}
\providecommand{\BIBentryALTinterwordstretchfactor}{4}
\providecommand{\BIBentryALTinterwordspacing}{\spaceskip=\fontdimen2\font plus
\BIBentryALTinterwordstretchfactor\fontdimen3\font minus
  \fontdimen4\font\relax}
\providecommand{\BIBforeignlanguage}[2]{{%
\expandafter\ifx\csname l@#1\endcsname\relax
\typeout{** WARNING: IEEEtran.bst: No hyphenation pattern has been}%
\typeout{** loaded for the language `#1'. Using the pattern for}%
\typeout{** the default language instead.}%
\else
\language=\csname l@#1\endcsname
\fi
#2}}
\providecommand{\BIBdecl}{\relax}
\BIBdecl

\bibitem{Abbasi_2016}
A.~Abbasi, ``A longitudinal analysis of link formation on collaboration
  networks,'' \emph{Journal of Informetrics}, vol.~10, no.~3, pp. 685--692,
  2016.

\bibitem{wang2017knowledge}
Q.~Wang, Z.~Mao, B.~Wang, and L.~Guo, ``Knowledge graph embedding: A survey of
  approaches and applications,'' \emph{IEEE Transactions on Knowledge and Data
  Engineering}, vol.~29, no.~12, pp. 2724--2743, 2017.

\bibitem{bengio2013representation}
Y.~Bengio, A.~Courville, and P.~Vincent, ``Representation learning: A review
  and new perspectives,'' \emph{IEEE transactions on pattern analysis and
  machine intelligence}, vol.~35, no.~8, pp. 1798--1828, 2013.

\bibitem{perozzi2014deepwalk}
B.~Perozzi, R.~Al-Rfou, and S.~Skiena, ``Deepwalk: Online learning of social
  representations,'' in \emph{Proceedings of the 20th ACM SIGKDD international
  conference on Knowledge discovery and data mining}.\hskip 1em plus 0.5em
  minus 0.4em\relax ACM, 2014, pp. 701--710.

\bibitem{tang2015line}
J.~Tang, M.~Qu, M.~Wang, M.~Zhang, J.~Yan, and Q.~Mei, ``Line: Large-scale
  information network embedding,'' in \emph{Proceedings of the 24th
  international conference on world wide web}.\hskip 1em plus 0.5em minus
  0.4em\relax International World Wide Web Conferences Steering Committee,
  2015, pp. 1067--1077.

\bibitem{cao2015grarep}
S.~Cao, W.~Lu, and Q.~Xu, ``Grarep: Learning graph representations with global
  structural information,'' in \emph{Proceedings of the 24th ACM international
  on conference on information and knowledge management}.\hskip 1em plus 0.5em
  minus 0.4em\relax ACM, 2015, pp. 891--900.

\bibitem{mahdavi2018dynnode2vec}
S.~Mahdavi, S.~Khoshraftar, and A.~An, ``dynnode2vec: Scalable dynamic network
  embedding,'' in \emph{2018 IEEE International Conference on Big Data (Big
  Data)}.\hskip 1em plus 0.5em minus 0.4em\relax IEEE, 2018, pp. 3762--3765.

\bibitem{haddad2019temporalnode2vec}
M.~Haddad, C.~Bothorel, P.~Lenca, and D.~Bedart, ``Temporalnode2vec: Temporal
  node embedding in temporal networks,'' in \emph{International Conference on
  Complex Networks and Their Applications}.\hskip 1em plus 0.5em minus
  0.4em\relax Springer, 2019, pp. 891--902.

\bibitem{singer2019node}
U.~Singer, I.~Guy, and K.~Radinsky, ``Node embedding over temporal graphs,''
  \emph{arXiv preprint arXiv:1903.08889}, 2019.

\bibitem{leskovec2005graphs}
J.~Leskovec, J.~Kleinberg, and C.~Faloutsos, ``Graphs over time: densification
  laws, shrinking diameters and possible explanations,'' in \emph{Proceedings
  of the eleventh ACM SIGKDD international conference on Knowledge discovery in
  data mining}.\hskip 1em plus 0.5em minus 0.4em\relax ACM, 2005, pp. 177--187.

\bibitem{Abbas2018Popularity}
K.~Abbas, M.~Shang, A.~Abbasi, X.~Luo, J.~J. Xu, and Y.~X. Zhang, ``Popularity
  and novelty dynamics in evolving networks,'' \emph{Scientific Reports},
  vol.~8, no.~1, 2018.

\bibitem{yu2016network}
F.~Yu, A.~Zeng, S.~Gillard, and M.~Medo, ``Network-based recommendation
  algorithms: A review,'' \emph{Physica A: Statistical Mechanics and its
  Applications}, vol. 452, pp. 192--208, 2016.

\bibitem{albert2002statistical}
R.~Albert and A.-L. Barab{\'a}si, ``Statistical mechanics of complex
  networks,'' \emph{Reviews of modern physics}, vol.~74, no.~1, p.~47, 2002.

\bibitem{trivedi2017know}
R.~Trivedi, H.~Dai, Y.~Wang, and L.~Song, ``Know-evolve: Deep temporal
  reasoning for dynamic knowledge graphs,'' in \emph{Proceedings of the 34th
  International Conference on Machine Learning-Volume 70}.\hskip 1em plus 0.5em
  minus 0.4em\relax JMLR. org, 2017, pp. 3462--3471.

\bibitem{lu2011link}
L.~L{\"u} and T.~Zhou, ``Link prediction in complex networks: A survey,''
  \emph{Physica A: statistical mechanics and its applications}, vol. 390,
  no.~6, pp. 1150--1170, 2011.

\bibitem{cui2018survey}
P.~Cui, X.~Wang, J.~Pei, and W.~Zhu, ``A survey on network embedding,''
  \emph{IEEE Transactions on Knowledge and Data Engineering}, 2018.

\bibitem{ou2016asymmetric}
M.~Ou, P.~Cui, J.~Pei, Z.~Zhang, and W.~Zhu, ``Asymmetric transitivity
  preserving graph embedding,'' in \emph{Proceedings of the 22nd ACM SIGKDD
  international conference on Knowledge discovery and data mining}.\hskip 1em
  plus 0.5em minus 0.4em\relax ACM, 2016, pp. 1105--1114.

\bibitem{cao2016deep}
S.~Cao, W.~Lu, and Q.~Xu, ``Deep neural networks for learning graph
  representations,'' in \emph{Thirtieth AAAI Conference on Artificial
  Intelligence}, 2016.

\bibitem{wang2016structural}
D.~Wang, P.~Cui, and W.~Zhu, ``Structural deep network embedding,'' in
  \emph{Proceedings of the 22nd ACM SIGKDD international conference on
  Knowledge discovery and data mining}.\hskip 1em plus 0.5em minus 0.4em\relax
  ACM, 2016, pp. 1225--1234.

\bibitem{kipf2016semi}
T.~N. Kipf and M.~Welling, ``Semi-supervised classification with graph
  convolutional networks,'' \emph{arXiv preprint arXiv:1609.02907}, 2016.

\bibitem{chen2018harp}
H.~Chen, B.~Perozzi, Y.~Hu, and S.~Skiena, ``Harp: Hierarchical representation
  learning for networks,'' in \emph{Thirty-Second AAAI Conference on Artificial
  Intelligence}, 2018.

\bibitem{grover2016node2vec}
A.~Grover and J.~Leskovec, ``node2vec: Scalable feature learning for
  networks,'' in \emph{Proceedings of the 22nd ACM SIGKDD international
  conference on Knowledge discovery and data mining}.\hskip 1em plus 0.5em
  minus 0.4em\relax ACM, 2016, pp. 855--864.

\bibitem{NguyenContinousTime2018}
\BIBentryALTinterwordspacing
G.~H. Nguyen, J.~B. Lee, R.~A. Rossi, N.~K. Ahmed, E.~Koh, and S.~Kim,
  ``Continuous-time dynamic network embeddings,'' in \emph{Companion
  Proceedings of the The Web Conference 2018}.\hskip 1em plus 0.5em minus
  0.4em\relax Republic and Canton of Geneva, CHE: International World Wide Web
  Conferences Steering Committee, 2018. [Online]. Available:
  \url{https://doi.org/10.1145/3184558.3191526}
\BIBentrySTDinterwordspacing

\bibitem{peng2019dynamic}
H.~Peng, J.~Li, H.~Yan, Q.~Gong, S.~Wang, L.~Liu, L.~Wang, and X.~Ren,
  ``Dynamic network embedding via incremental skip-gram with negative
  sampling,'' \emph{arXiv preprint arXiv:1906.03586}, 2019.

\bibitem{li2018deep}
T.~Li, J.~Zhang, S.~Y. Philip, Y.~Zhang, and Y.~Yan, ``Deep dynamic network
  embedding for link prediction,'' \emph{IEEE Access}, vol.~6, pp.
  29\,219--29\,230, 2018.

\bibitem{demmel1990matrix}
J.~W. Demmel, ``Matrix computations; (gene golub and charles f. van loan),''
  \emph{SIAM Review}, vol.~32, no.~4, p. 690, 1990.

\bibitem{cho2014learning}
K.~Cho, B.~Van~Merri{\"e}nboer, C.~Gulcehre, D.~Bahdanau, F.~Bougares,
  H.~Schwenk, and Y.~Bengio, ``Learning phrase representations using rnn
  encoder-decoder for statistical machine translation,'' \emph{arXiv preprint
  arXiv:1406.1078}, 2014.

\bibitem{kingma2014adam}
D.~P. Kingma and J.~Ba, ``Adam: A method for stochastic optimization,''
  \emph{arXiv preprint arXiv:1412.6980}, 2014.

\bibitem{ijcai2019ProNE}
\BIBentryALTinterwordspacing
J.~Zhang, Y.~Dong, Y.~Wang, J.~Tang, and M.~Ding, ``Prone: Fast and scalable
  network representation learning,'' in \emph{Proceedings of the Twenty-Eighth
  International Joint Conference on Artificial Intelligence, {IJCAI-19}}.\hskip
  1em plus 0.5em minus 0.4em\relax International Joint Conferences on
  Artificial Intelligence Organization, 7 2019, pp. 4278--4284. [Online].
  Available: \url{https://doi.org/10.24963/ijcai.2019/594}
\BIBentrySTDinterwordspacing

\bibitem{mikolov2013efficient}
T.~Mikolov, K.~Chen, G.~Corrado, and J.~Dean, ``Efficient estimation of word
  representations in vector space,'' \emph{arXiv preprint arXiv:1301.3781},
  2013.

\bibitem{realityMiningMIT2015Download}
\BIBentryALTinterwordspacing
``Reality mining network dataset -- {KONECT},'' Apr. 2015. [Online]. Available:
  \url{http://konect.uni-koblenz.de/networks/mit}
\BIBentrySTDinterwordspacing

\bibitem{realityMiningMIT2015}
N.~Eagle and A.~(Sandy)~Pentland, ``{Reality} {Mining}: Sensing complex social
  systems,'' \emph{Personal Ubiquitous Computing}, vol.~10, no.~4, pp.
  255--268, 2006.

\bibitem{nr-aaai15}
\BIBentryALTinterwordspacing
R.~A. Rossi and N.~K. Ahmed, ``The network data repository with interactive
  graph analytics and visualization,'' in \emph{AAAI}, 2015. [Online].
  Available: \url{http://networkrepository.com}
\BIBentrySTDinterwordspacing

\end{thebibliography}

\end{document}